\documentclass{article}
\usepackage{spconf,amsmath,graphicx}
\usepackage{color}
\usepackage{booktabs} 
\usepackage{multirow} 
\usepackage{float}
\usepackage{bm}
\usepackage[table]{xcolor}
\usepackage{subcaption}
\usepackage{amssymb}
\usepackage{url}

\title{Leveraging Vision-Language Models as Weak Annotators \\in Active Learning}

\name{Phuong Ngoc Nguyen, Kaito Shiku, Ryoma Bise, Seiichi Uchida, Shinnosuke Matsuo}
\address{\vspace{-8mm}\\
Kyushu University, Japan.\\
{\tt\small \{ngoc.phuong,shinnosuke.matsuo\}@human.ait.kyushu-u.ac.jp}
}

\begin{document}
\ninept
\maketitle

\begin{abstract}
Active learning aims to reduce annotation cost by selectively querying informative samples for supervision under a limited labeling budget. In this work, we investigate how vision-language models (VLMs) can be leveraged to further reduce the reliance on costly human annotation within the active learning paradigm. To this end, we find that the reliability of VLMs varies significantly with label granularity in fine-grained recognition tasks: they perform poorly on fine-grained labels but can provide accurate coarse-grained labels. Leveraging this property, we propose an active learning framework that combines fine-grained human annotations with coarse-grained VLM-generated weak labels through instance-wise label assignment. We further model the systematic noise in VLM-generated labels using a small set of trusted full labels. Experiments on CUB200 and FGVC-Aircraft show that the proposed framework consistently outperforms existing active learning methods under the same annotation budget.
\end{abstract}
\begin{keywords}
Active learning, Weak supervision, Vision-language models
\end{keywords}

\section{Introduction}
\label{sec:intro}

Active learning (AL)~\cite{settles2009active,ren2021survey,aggarwal2014active, wang2014new, ashdeep, Matsuo_2025_CVPR} aims to improve model performance under a limited annotation budget by selectively querying the most informative data samples for annotation.
A common strategy is to prioritize samples near the decision boundary, where additional supervision is expected to yield the largest performance gain.
In conventional AL frameworks, annotations are typically assumed to be provided by human annotators, often domain experts, who supply accurate but costly labels.

To alleviate the cost of human annotation, recent advances in vision-language models (VLMs), such as large-scale pretrained multimodal models~\cite{radford2021learning, google2024gemini2flash, bang2024active, safaei2025active}, have opened up the possibility of using them as alternative annotators.
Modern VLMs possess broad and general visual knowledge acquired from large-scale pretraining, enabling them to assign labels for common concepts without additional supervision.
However, for fine-grained classification tasks, such as distinguishing bird species or aircraft variants, accurate labeling requires specialized domain knowledge, making it difficult for VLMs to provide reliable fine-grained labels.

Although VLMs are not well suited for assigning fine-grained labels that require specialized domain knowledge, we find that they can reliably assign coarse labels corresponding to higher-level categories in the class hierarchy. For example, while distinguishing between visually similar bird species, such as ``Song Sparrow'' and ``House Sparrow'', is challenging for VLMs, assigning a coarse label, such as ``Sparrow'', can be achieved with much higher accuracy.
This observation motivates the central idea of this work and is empirically supported by our preliminary experiments.
On the CUB200 dataset~\cite{WahCUB_200_2011}, VLM-based annotation achieves only 67.08\% accuracy for fine-grained full labels, whereas it reaches 85.24\% accuracy for coarse-grained weak labels. 
A similar trend is observed on the FGVC-Aircraft dataset~\cite{maji2013fine}, where full-label accuracy is 59.26\%, while weak-label accuracy improves to 82.11\%.

\begin{figure*}[t] 
    \centering
    \includegraphics[width=.78\linewidth]{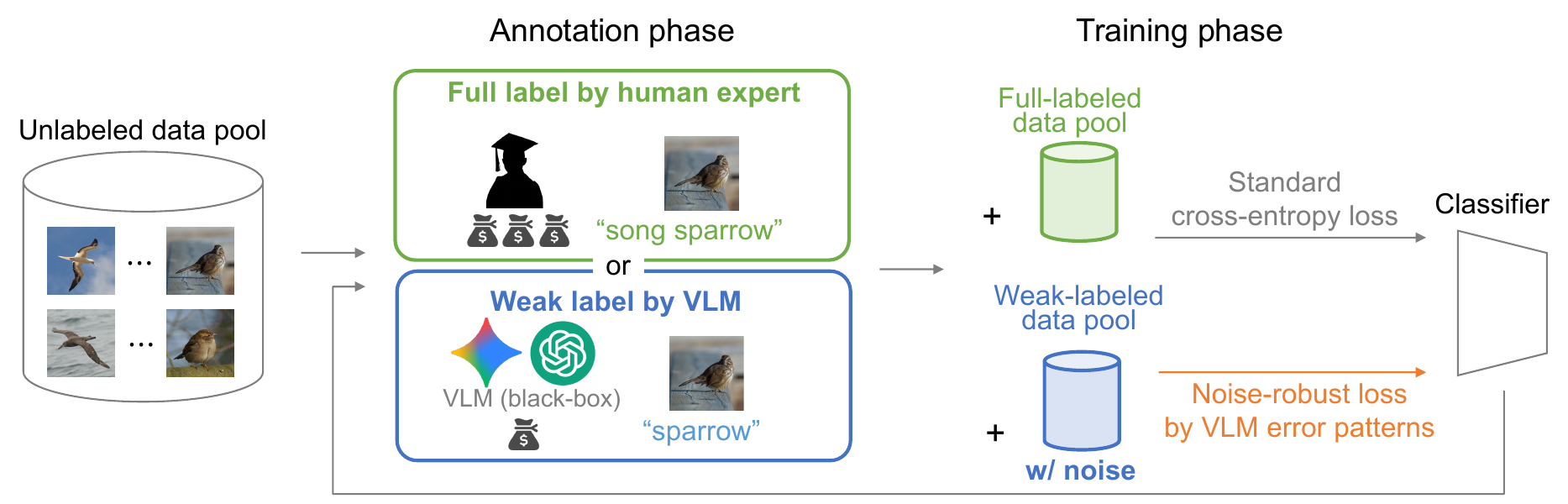} 
    \caption{Overview of the proposed method.
The proposed active learning framework integrates a vision-language model (VLM) as a weak annotator, allowing each queried instance to be annotated either with a fine-grained human label or a coarse-grained VLM-generated weak label under a fixed annotation budget.
Weakly labeled samples are trained using a noise-robust loss that accounts for VLM-specific error patterns.}

    \label{fig:overview}
\end{figure*}

Based on this observation, we propose a new active learning framework that leverages VLMs as weak annotators.
The proposed framework combines accurate but expensive full labels provided by human experts with inexpensive but noisy weak labels generated by VLMs, as illustrated in Fig.~\ref{fig:overview}.
Importantly, our framework explicitly models the annotation costs and label granularities associated with human experts and VLMs, viewing them as heterogeneous supervision sources with different cost–reliability characteristics.
Under this formulation, active learning is cast as the problem of assigning, to each data instance, either a high-cost fine-grained label from a human expert or a low-cost coarse-grained (weak) label generated by a VLM, subject to a fixed total annotation budget.

Since weak labels generated by VLMs often contain noise, we further incorporate a noise-robust learning strategy that explicitly models how VLMs make systematic errors.
Specifically, we estimate how VLM predictions deviate from ground-truth labels using a small set of trusted full labels provided by human experts, and encode this error structure in a noise transition matrix.
This transition matrix is then used to correct the influence of noisy weak labels during training through a noise-robust loss function, leading to more stable learning from VLM-generated supervision.

We evaluate the proposed method on fine-grained classification benchmarks, including CUB200 and FGVC-Aircraft.
Under the same total annotation cost, our approach consistently outperforms conventional active learning methods that rely solely on human-provided full labels.
Moreover, ablation studies demonstrate that incorporating learning-with-noisy-labels techniques significantly improves performance, confirming their effectiveness in reducing noise introduced by VLM-based weak annotation.

 Our main contributions are summarized as follows:
\begin{itemize}
    \item We find that the reliability of VLMs varies significantly with label granularity in fine-grained recognition, showing that VLMs are unreliable for fine-grained labels but can provide accurate coarse-grained (weak) labels.
    \item Leveraging this observation, we propose an active learning framework that combines fine-grained human annotations with coarse-grained VLM-generated weak labels through instance-wise label assignment.
    \item We explicitly model the systematic noise in VLM-generated weak labels using a small set of trusted full labels, enabling robust learning; experiments on fine-grained benchmarks demonstrate consistent improvements over conventional active learning under the same annotation budget.
\end{itemize}

\section{Related Work}
\label{sec:related}

\subsection{Active Learning}

Active learning (AL)~\cite{settles2009active,ren2021survey,aggarwal2014active} aims to improve model performance under a limited annotation budget by selectively querying informative samples for labeling.
Many AL methods select samples based on predictive uncertainty, prioritizing instances that are expected to provide the largest performance gain, using criteria such as entropy~\cite{wang2014new}, margin-based measures, and Bayesian uncertainty estimation~\cite{gal2017deep}.
Another important line of work focuses on diversity sampling, which aims to avoid redundant queries by selecting a representative and diverse subset of samples.
Representative approaches include core-set selection~\cite{sener2018active} and BADGE~\cite{ashdeep}.
BADGE jointly considers uncertainty and diversity in the gradient embedding space.
AL has also been studied in settings where multiple levels or granularities of supervision are available~\cite{Matsuo_2025_CVPR,tejero2023full}, in which the type or level of supervision is selected for each instance under a given annotation budget, for example by considering annotation cost~\cite{Matsuo_2025_CVPR}.

Existing active learning frameworks typically rely on human annotators to provide supervision.
In this work, we incorporate vision-language models (VLMs) as an additional source of weak supervision to reduce annotation cost, while retaining accurate fine-grained labels from human experts.

\subsection{Active Learning for Vision-Language Models}

Recent studies have explored active learning for adapting vision-language models (VLMs), particularly CLIP~\cite{radford2021learning}, to downstream tasks~\cite{bang2024active, safaei2025active}.
In this line of work, the VLM itself serves as the target classifier, and active learning is used to improve its performance through mechanisms such as prompt tuning or calibration.
For example, Bang et al.~\cite{bang2024active} apply active learning to prompt tuning in VLMs, addressing the tendency of zero-shot CLIP to collapse to a small subset of classes by performing class-balanced sampling based on pseudo-labels.

In contrast to these approaches, which aim to adapt or improve VLMs themselves, our work focuses on \emph{VLMs as weak annotators}.
Specifically, we consider large-scale vision-language models~\cite{google2024gemini2flash, achiam2023gpt} that are accessed as black boxes and use them solely as external sources of coarse-grained supervision in active learning, while training a separate visual classifier.
This design choice is motivated by the broader visual knowledge provided by such models compared to standalone VLMs.

\section{Preliminary Experiments}
\label{sec:preliminary}

To use vision-language models (VLMs) as annotators in active learning for fine-grained classification, we first examine how their inference performance depends on label granularity.
Specifically, we compare VLM performance on fine-grained and coarse-grained class labels.

\subsection{Datasets}
We conduct preliminary experiments on two fine-grained classification benchmarks.

\par\noindent\textbf{1) Caltech-UCSD Birds-200-2011 (CUB200)~\cite{WahCUB_200_2011}:}
A bird classification dataset containing 11,788 images across 200 species.
In our study, species-level labels are treated as fine-grained labels.
Coarse-grained labels are defined following~\cite{Lu2018UsingCL}, where 70 superclasses are constructed based on suffix patterns in the class names.
The dataset consists of 5,994 training images and 5,794 evaluation images.

\par\noindent\textbf{2) FGVC-Aircraft~\cite{maji2013fine}:}
An aircraft classification dataset consisting of 10,000 images covering 100 variants.
We use the coarse-grained labels as predefined in the dataset, where the coarse-grained label corresponds to the manufacturer level.
The dataset consists of 6,667 training images and 3,333 evaluation images.


\subsection{Experimental settings}
We adopt Gemini 2.0 Flash~\cite{google2024gemini2flash}, provided as a paid commercial service API, as the vision-language model (VLM).
For inference, we use the prompt:\textit{``Classify the image into one of the following $N$ classes: CLASS$_1$, \dots, CLASS$_N$.''}
Here, \textit{CLASS$_1$, \dots, CLASS$_N$} correspond to the class names.
If the output cannot be mapped to any candidate class, we treat it as an invalid response and exclude it from evaluation.
When evaluating fine-grained classification capability, fine-grained class names are used in the prompt, whereas coarse-grained class names corresponding to higher-level categories are used when evaluating coarse-grained classification performance.

\begin{table}[t] 
    \centering
    \caption{Performance comparison of Gemini on Fine-grained vs. Coarse Label [\%].}
    \label{tab:prelim_results}
  \scalebox{1.0}{
    \begin{tabular}{lcc}
        \toprule
        \text{Class} & CUB200~\cite{WahCUB_200_2011} & FGVC-Aircraft~\cite{maji2013fine} \\
        \midrule
        Fine-grained & 67.08& 59.26\\ 
        Coarse-grained & \textbf{85.24} & \textbf{82.11} \\
        \bottomrule
    \end{tabular}
    }
\end{table}

\subsection{Performance evaluation}
Table~\ref{tab:prelim_results} presents the prediction results obtained using fine-grained and coarse-grained labels on the CUB200 and FGVC-Aircraft datasets. 
These results reveal that fine-grained classification capability is limited on both datasets, whereas coarse-grained classification performance is substantially higher.

These observations suggest that, while VLMs possess extensive general knowledge that enables them to infer coarse-grained classes, they lack sufficient domain-specific knowledge to predict fine-grained labels in these tasks accurately.
Based on the results of these preliminary experiments, the proposed method leverages VLMs as low-cost coarse-grained annotators.

\section{Proposed Method}
\label{sec:method}
\subsection{Problem Setting and Overview}

We consider an active learning setting for fine-grained image classification under a limited annotation budget, where each queried instance is annotated either with a fine-grained human label (full label) or a coarse-grained label (weak label) generated by a vision-language model (VLM), as illustrated in Fig.~\ref{fig:overview}.
Human labels are accurate but costly, whereas VLM-generated labels are inexpensive but coarse and potentially noisy.
We assume black-box access to a large-scale VLM, which is used solely as a source of weak supervision to train a standalone visual classifier.

Formally, let $D_u$ denote the unlabeled data pool and $D_I$ a small set of initially labeled instances with fine-grained human annotations.
A classifier $f$ is trained on $D_I$.
The active learning process proceeds over $T$ rounds: at each round, candidate instances are selected from $D_u$ based on the outputs of $f$ (e.g., uncertainty and diversity criteria), annotated with either a human label with cost $C_f$ or a VLM-generated label with cost $C_w$ ($C_f \gg C_w$), and used to update $f$.
The objective is to maximize fine-grained classification accuracy after $T$ rounds under the budget constraint.

\begin{table*}[t] 
    \centering
    \caption{Comparison with the conventional AL methods on CUB200. This table shows the classification accuracy [\%].}

    \label{tab:comparison_cub200} 
    \vspace{-2mm}
    
    \setlength{\tabcolsep}{10pt}
  \scalebox{1.0}{
    \begin{tabular}{l c c c c c c c}
        \toprule
        \multicolumn{1}{c}{} 
        & \multicolumn{2}{c}{\textbf{Annotator}} 
        & \multicolumn{5}{c}{\textbf{Round}} \\
        \cmidrule(lr){2-3}
        \cmidrule(lr){4-8}
        \textbf{Method} 
        & \textbf{Human} 
        & \textbf{VLM} 
        & \textbf{1} 
        & \textbf{2} 
        & \textbf{3} 
        & \textbf{4} 
        & \textbf{5} \\
        \midrule
        Random 
        & \checkmark & -- 
        & 34.41 $\pm$ 4.82 
        & 41.89 $\pm$ 4.17 
        & 47.99 $\pm$ 3.53 
        & 53.51 $\pm$ 3.13 
        & 58.09 $\pm$ 2.23 \\
        
        Entropy~\cite{wang2014new} 
        & \checkmark & -- 
        & 34.73 $\pm$ 5.05 
        & 43.19 $\pm$ 4.18 
        & 50.79 $\pm$ 3.21 
        & 56.77 $\pm$ 3.06 
        & 60.93 $\pm$ 2.42 \\
        
        BADGE~\cite{ashdeep}   
        & \checkmark & -- 
        & 34.24 $\pm$ 4.10 
        & 41.87 $\pm$ 3.63 
        & 48.57 $\pm$ 3.06 
        & 55.96 $\pm$ 2.63 
        & 60.28 $\pm$ 2.46 \\
        
        ISOAL~\cite{Matsuo_2025_CVPR}  
        & \checkmark & -- 
        & 58.86 $\pm$ 1.64 
        & 61.81 $\pm$ 2.22 
        & 66.74 $\pm$ 2.23 
        & 67.98 $\pm$ 1.07 
        & 71.37 $\pm$ 1.03 \\
        
        \midrule
        \rowcolor{gray!15} 
        {\bf Ours}  
        & \checkmark & \checkmark 
        & \textbf{59.98 $\pm$ 1.47}
        & \textbf{62.94 $\pm$ 2.04}  
        & \textbf{67.55 $\pm$ 1.14}  
        & \textbf{70.58 $\pm$ 1.66}  
        & \textbf{71.67 $\pm$ 0.81} \\
        

        \bottomrule
    \end{tabular}
    }
    \centering
    \vspace{6mm}
    \caption{Comparison with the conventional AL methods on FGVC-Aircraft. This table shows the classification accuracy [\%].}
    \label{tab:comparison_aircraft} 
    \vspace{-2mm}
  \scalebox{1.0}{
    \begin{tabular}{l c c c c c c c}
        \toprule
        \multicolumn{1}{c}{} 
        & \multicolumn{2}{c}{\textbf{Annotator}} 
        & \multicolumn{5}{c}{\textbf{Round}} \\
        \cmidrule(lr){2-3}
        \cmidrule(lr){4-8}
        \textbf{Method} 
        & \textbf{Human} 
        & \textbf{VLM} 
        & \textbf{1} 
        & \textbf{2} 
        & \textbf{3} 
        & \textbf{4} 
        & \textbf{5} \\
        \midrule
        Random  
        & \checkmark & -- 
        & 14.04 $\pm$ 1.41 
        & 18.47 $\pm$ 1.50 
        & 22.86 $\pm$ 2.37 
        & 27.07 $\pm$ 1.93 
        & 30.84 $\pm$ 2.42 \\
        
        Entropy~\cite{wang2014new} 
        & \checkmark & -- 
        & 13.23 $\pm$ 1.76 
        & 17.05 $\pm$ 1.66 
        & 20.03 $\pm$ 1.82 
        & 23.49 $\pm$ 1.09 
        & 26.93 $\pm$ 1.32 \\
        
        BADGE~\cite{ashdeep}   
        & \checkmark & -- 
        & 13.34 $\pm$ 1.47 
        & 17.37 $\pm$ 1.23 
        & 22.06 $\pm$ 1.67 
        & 25.24 $\pm$ 2.05 
        & 29.11 $\pm$ 1.96 \\
        
        ISOAL~\cite{Matsuo_2025_CVPR} 
        & \checkmark & -- 
        & 27.73 $\pm$ 3.73  
        & 39.56 $\pm$ 3.16  
        & 45.29 $\pm$ 1.23  
        & 49.84 $\pm$ 0.61  
        & 54.45 $\pm$ 0.87 \\
        
        \midrule
        \rowcolor{gray!15} 
        {\bf Ours}  
        & \checkmark & \checkmark 
        & \textbf{29.63 $\pm$ 8.23}  
        & \textbf{43.78 $\pm$ 7.02}  
        & \textbf{50.22 $\pm$ 1.72}  
        & \textbf{54.56 $\pm$ 1.11}  
        & \textbf{57.78 $\pm$ 0.35} \\
        
        \bottomrule
    \end{tabular}
    }
    \vspace{2mm}
\end{table*}

\begin{table*}[t]
    \centering
    \caption{Ablation study on CUB200. This table shows the classification accuracy [\%].}
    \label{tab:ablation_cub200} 
    \vspace{-2mm}
    \setlength{\tabcolsep}{10pt}
    \begin{tabular}{l c c c c c}
        \toprule
        \multicolumn{1}{c}{} & \multicolumn{5}{c}{\textbf{Round}} \\
        \cmidrule(lr){2-6}
        \textbf{Method} 
        & \textbf{1} 
        & \textbf{2} 
        & \textbf{3} 
        & \textbf{4} 
        & \textbf{5} \\
        \midrule
        \rowcolor{gray!15} 
        {\bf Ours}  
        & 59.98 $\pm$ 1.47  
        & 62.94 $\pm$ 2.04
        & \textbf{67.55 $\pm$ 1.14}  
        & \textbf{70.58 $\pm$ 1.66}  
        & \textbf{71.67 $\pm$ 0.81} \\
        
        \ \ w/o Noise-robust loss $\mathcal{L}_{\text{weak}}$
        & \textbf{61.87 $\pm$ 1.01}
        & \textbf{63.62 $\pm$ 1.62}
        & 67.40 $\pm$ 1.92 
        & 69.49 $\pm$ 1.47 
        & 70.95 $\pm$ 1.44 \\
        \bottomrule
    \end{tabular}
    \centering
    \vspace{6mm}
    \caption{Ablation study on FGVC-Aircraft. This table shows the classification accuracy [\%].}
    \label{tab:ablation_aircraft} 
    \vspace{-2mm}
    \setlength{\tabcolsep}{10pt}
    \begin{tabular}{l c c c c c}
        \toprule
        \multicolumn{1}{c}{} & \multicolumn{5}{c}{\textbf{Round}} \\
        \cmidrule(lr){2-6}
        \textbf{Method} 
        & \textbf{1} 
        & \textbf{2} 
        & \textbf{3} 
        & \textbf{4} 
        & \textbf{5} \\
        \midrule
        \rowcolor{gray!15} 
        {\bf Ours}  
        & 29.63 $\pm$ 8.23  
        & 43.78 $\pm$ 7.02  
        & \textbf{50.22 $\pm$ 1.72}  
        & \textbf{54.56 $\pm$ 1.11}  
        & \textbf{57.78 $\pm$ 0.35} \\
        
        \ \ w/o Noise-robust loss $\mathcal{L}_{\text{weak}}$
        & \textbf{32.23 $\pm$ 6.96}  
        & \textbf{45.18 $\pm$ 0.57}  
        & 49.39 $\pm$ 2.09  
        & 53.74 $\pm$ 0.67  
        & 56.93 $\pm$ 1.35 \\
        \bottomrule
    \end{tabular}

    \centering
    \vspace{6mm}
    \caption{Classification accuracy [\%] with varying weak annotation costs of VLM on CUB200.}
    \label{tab:weak_cost} 
    \vspace{-2mm}
    \setlength{\tabcolsep}{10pt}
    \begin{tabular}{l c c c c c}
        \toprule
        \multicolumn{1}{c}{} & \multicolumn{5}{c}{\textbf{Round}} \\
        \cmidrule(lr){2-6}
        \textbf{Method} 
        & \textbf{1} 
        & \textbf{2} 
        & \textbf{3} 
        & \textbf{4} 
        & \textbf{5} \\
        \midrule
        {\bf Ours} \ ($C_w = 1/20$)  
        & 60.34 $\pm$ 1.60 
        & 62.95 $\pm$ 0.76
        & 67.03 $\pm$ 1.58 
        & 68.07 $\pm$ 0.63 
        & 72.02 $\pm$ 1.09 \\ 
        
        {\bf Ours} \ ($C_w = 1/50$)  
        & 59.98 $\pm$ 1.47 
        & 62.94 $\pm$ 2.04 
        & 67.55 $\pm$ 1.14 
        & 70.58 $\pm$ 1.66 
        & 71.67 $\pm$ 0.81  \\ 
        
        {\bf Ours} \ ($C_w = 1/100$)  
        & 62.18 $\pm$ 0.71 
        & 65.94 $\pm$ 0.54 
        & 68.36 $\pm$ 2.22 
        & 71.17 $\pm$ 1.79 
        & 71.26 $\pm$ 2.17 \\ 
        \bottomrule
    \end{tabular}
\end{table*}

\subsection{Active Learning with Human and VLM Supervision}

The proposed method integrates weak labels generated by a vision-language model (VLM) into the active learning loop.
At each round, a subset of instances is selected from the unlabeled data pool $D_u$, and for each queried instance, the learner assigns the appropriate supervision source (human or VLM) by accounting for their different annotation costs under a limited budget.

To decide, under a fixed annotation budget, which instances in $D_u$ to query and whether to obtain an expensive fine-grained human label or a low-cost coarse-grained VLM label, we use an off-the-shelf instance-wise supervision allocation solver~\cite{Matsuo_2025_CVPR} as a joint selection-and-assignment module.
This allocation component is interchangeable, and the proposed framework does not rely on the particular choice of allocator.

Our weak supervision is provided by a black-box VLM and can exhibit systematic, model-specific errors.
As a result, treating VLM outputs as standard weak annotations may introduce label noise.
We therefore estimate the VLM's weak-label error patterns from a small trusted set and train the classifier with a noise-corrected objective, as described in the next section.

\subsection{Noise-Robust Learning with Forward Correction}

We handle noisy weak supervision from vision-language models (VLMs) using forward correction~\cite{patrini2017making}.
A small set of fine-grained human annotations is available from the beginning of active learning; for these samples, ground-truth weak labels can be derived and VLM zero-shot predictions can be obtained.
This allows us to estimate how the VLM misclassifies weak labels.

Let $K_w$ be the number of weak-label classes.
We define a VLM error transition matrix $T \in \mathbb{R}^{K_w \times K_w}$, where $T_{ij}$ denotes the probability that a sample with ground-truth weak label $i$ is predicted as $j$ by the VLM.
The matrix $T$ is estimated by comparing ground-truth weak labels (derived from fine-grained labels) with VLM zero-shot predictions on the initial labeled set.

Let $f_w(x) \in \mathbb{R}^{K_w}$ denote the predicted probability over weak classes.
For weakly labeled samples with observed label $y_w$, we use the forward correction loss
\begin{equation}
\mathcal{L}_{\text{weak}}(x, y_w) = \ell\!\left(T^\top f_w(x),\, y_w\right),
\end{equation}
where $\ell(\cdot,\cdot)$ denotes the standard cross-entropy loss.
Fine-grained human-labeled samples are trained using the standard cross-entropy loss.

During retraining, we first learn from abundant weak labels using the corrected loss, then refine the classifier with fine-grained human annotations.
This corrects systematic VLM errors while effectively leveraging weak supervision.

\section{Experiments}

\subsection{Dataset}
In this section, we use the same datasets as those employed in the preliminary experiments, namely CUB200~\cite{WahCUB_200_2011} and FGVC-Aircraft~\cite{maji2013fine}.
Both datasets are fine-grained image classification benchmarks with predefined hierarchical label structures, which allow us to naturally use fine-grained full labels and coarse-grained weak labels.

\subsection{Comparison methods}
We compare the proposed method with four representative AL methods that rely solely on conventional manual annotations and do not incorporate weak supervision from vision-language models.

\par\noindent\textbf{1) Random:}
A baseline that randomly samples instances from the unlabeled pool.

\par\noindent\textbf{2) Entropy~\cite{wang2014new}:}
An uncertainty-based AL method that selects instances with high entropy of the class probability predictions produced by the classifier.

\par\noindent\textbf{3) BADGE~\cite{ashdeep}:}
An AL method that accounts for both predictive uncertainty and diversity by selecting instances with diverse and high-magnitude gradients in the gradient space.

\par\noindent\textbf{4) ISOAL~\cite{Matsuo_2025_CVPR}:}
An AL framework that selects the supervision level for each instance under a fixed annotation budget, assuming all annotations are provided by human annotators.

\subsection{Implementation Details}
The proposed network architecture consists of a shared feature extractor and two classification heads, one for fully supervised learning and the other for weakly supervised learning.
We adopt Vision Transformer (ViT-B/16)~\cite{VIT} as the feature extractor, while each classification head is implemented as a single linear layer.
The network was optimized using the Adam~\cite{kingma2014adam} optimizer with a learning rate of 3e-05.
Each round was trained for 50 epochs, and the best model is selected based on the loss on the validation data.
Here, the validation data are constructed as a fully supervised set by sampling two instances from each class.
The performance of each method is evaluated by averaging the results over three different random seeds.

The settings for AL were as follows. The number of acquisition rounds $T$ was set to 5. The budget of each round was set to $150$ for both the CUB200 and FGVC-Aircraft datasets. 
The initial labeled set $D_I$ was constructed by randomly sampling three instances from each class.
The cost of fully supervised labels $C_f$ was set to 1, while the cost of weakly supervised labels $C_w$ was set to $1/50$, reflecting the significantly lower cost of labels generated by VLMs.
As in the preliminary experiments, we employed Gemini 2.0 Flash~\cite{google2024gemini2flash}, accessed via the Google DeepMind API, as the VLM.
In addition, for ISOAL, the cost of weak annotations provided by human annotators was set to $1/10$.
Note that these annotation costs are predefined based on practical considerations, such as annotation rewards and VLM API usage fees, rather than being treated as tunable hyperparameters.

\subsection{Comparison Experiments}
Tables~\ref{tab:comparison_cub200} and~\ref{tab:comparison_aircraft} present comparison results between the proposed method and conventional AL methods on CUB200 and FGVC-Aircraft, respectively.
For all methods, the annotation cost at each round is kept identical, and the reported results are presented as the mean and standard deviation over three runs with different random seeds.

As shown in the tables, the proposed method consistently outperforms conventional methods across all five rounds on both datasets.
While conventional AL methods assume that annotations are provided solely by human experts, the proposed method effectively leverages both human experts and VLMs.
Specifically, the proposed method introduces a VLM as a weak annotator and allocates a portion of the annotation cost to low-cost weak labels generated by the VLM instead of expensive full labels, enabling the use of a larger amount of labeled data.
By appropriately balancing fine-grained full labels provided by human experts and coarse-grained weak labels provided by the VLM, the proposed method achieves more efficient learning under the same annotation budget, leading to improved performance.

\subsection{Ablation Study}
Tables~\ref{tab:ablation_cub200} and~\ref{tab:ablation_aircraft} show ablation results on CUB200 and FGVC-Aircraft to evaluate the effectiveness of the noise-robust loss, which is designed to mitigate the impact of noise and systematic errors potentially contained in weak labels generated by the VLM.
We compare the proposed method with and without the noise-robust loss under the same experimental settings.

On both datasets, the effectiveness of the noise-robust loss is observed from Round~3 onward.
In the early rounds, when the amount of weakly labeled data is limited, the benefit of noise-robust learning is less evident.
As active learning progresses and a larger portion of the training set consists of VLM-generated weak labels, explicitly accounting for label noise becomes increasingly important.
These results indicate that the proposed noise-robust loss plays a crucial role in stabilizing learning from noisy weak supervision in later rounds.

\subsection{Cost Sensitivity Analysis of Weak Supervision}

To demonstrate that the proposed method operates effectively under different weak annotation costs of the VLM, Table~\ref{tab:weak_cost} presents experimental results on CUB200 with $C_w = 1/20$, $1/50$, and $1/100$.
In all experiments, the cost of fully supervised labels is set to $C_f = 1$, and the total annotation budget per round is fixed.

As shown in the table, the proposed method consistently achieves strong performance across all weak annotation cost settings.
In particular, performance improvements are observed not only at $C_w = 1/50$, but also at $C_w = 1/20$ and $C_w = 1/100$, indicating that the proposed framework is effective over a wide range of weak annotation costs.
This suggests that the proposed method is robust to the weak annotation cost and can operate effectively under various cost conditions.

Moreover, a general trend can be observed in which classification accuracy tends to improve as the weak annotation cost decreases.
This is because lower annotation costs allow a larger number of samples to be weakly annotated under a fixed budget, enabling the model to benefit from more training data provided by cost-efficient VLM-based supervision.

\begin{figure}[t]
    \centering
    \begin{subfigure}[b]{0.49\linewidth}
        \centering
        \includegraphics[width=\linewidth]{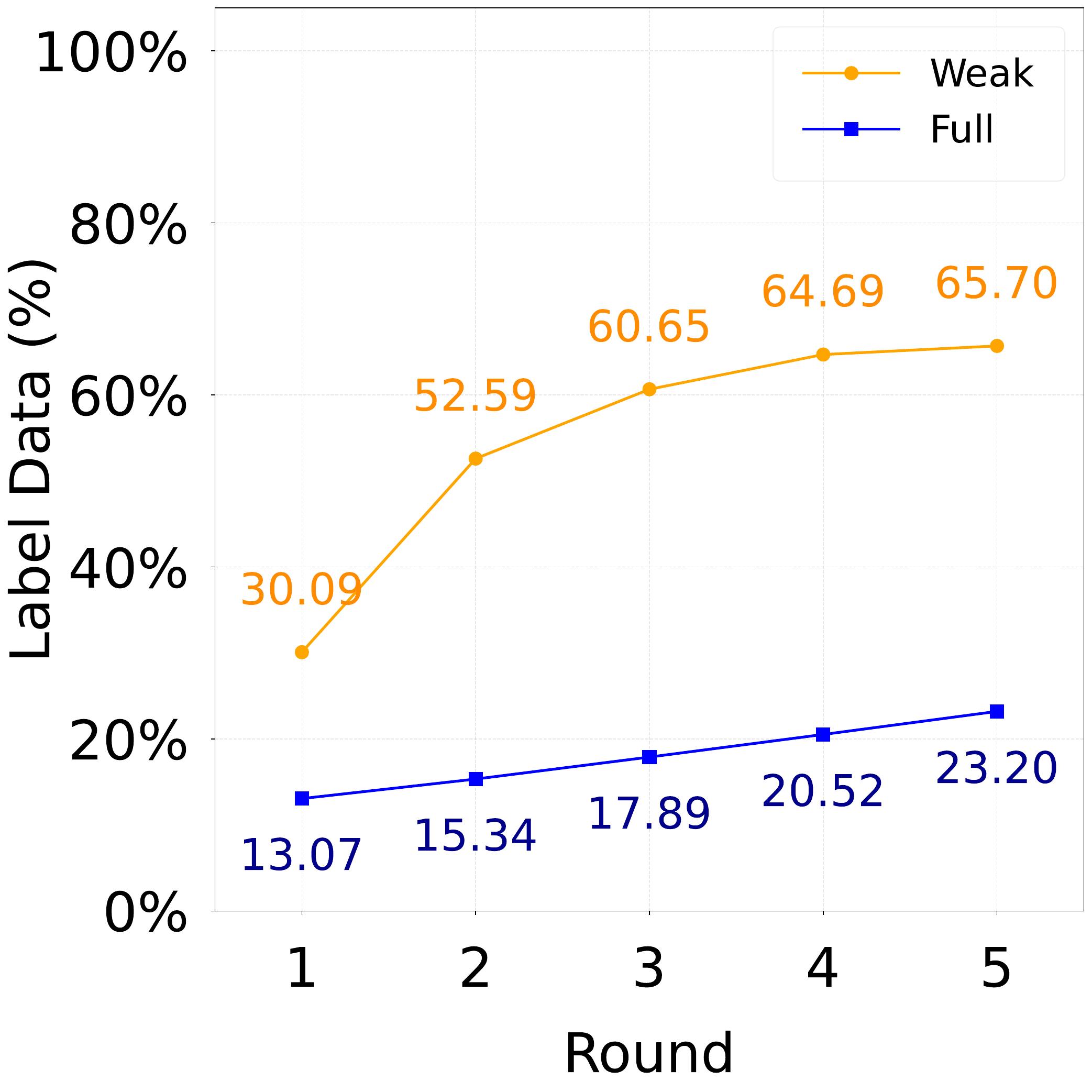}
        \caption{CUB200}
    \end{subfigure}
    \hfill
    \begin{subfigure}[b]{0.49\linewidth}
        \centering
        \includegraphics[width=\linewidth]{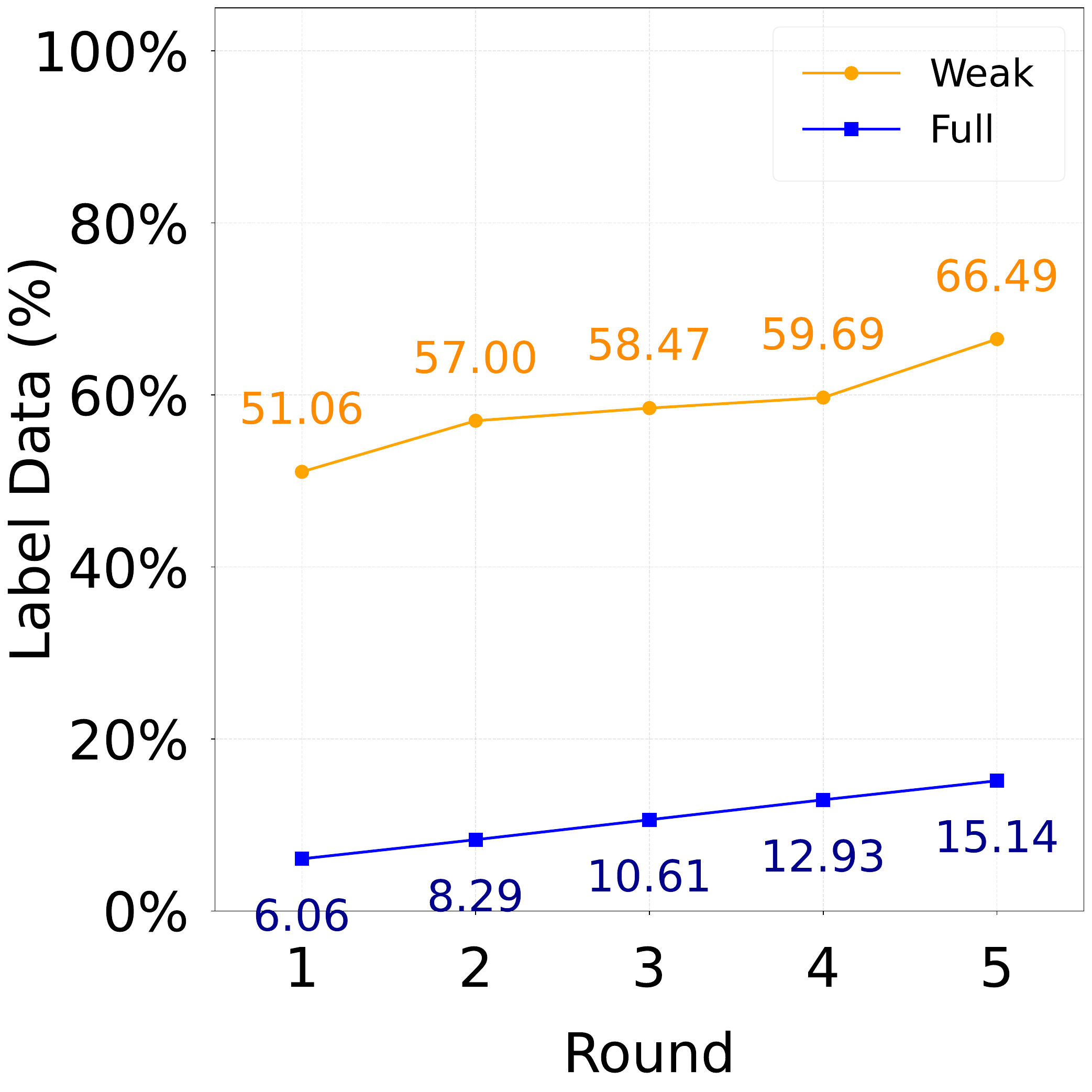}
        \caption{FGVC-Aircraft}
    \end{subfigure}
    \caption{Ratio of fully labeled data and weakly labeled data within the training dataset in the proposed method.}

    \label{fig:ratio}
\end{figure}

\subsection{Analysis of Labeled Data Ratio}

Fig.~\ref{fig:ratio} illustrates the ratio of fully labeled data by human experts and weakly labeled data by the VLM within the training dataset in the proposed method on CUB200 and FGVC-Aircraft.
On both datasets, the proposed method effectively utilizes a portion of the annotation budget for weak labeling by the VLM, rather than relying exclusively on costly fully supervised labels.

Through this cost allocation strategy, the proposed method can effectively exploit weak annotations and achieve improved performance compared to conventional active learning methods that rely solely on fully labeled data, as shown in Tables~\ref{tab:comparison_cub200} and~\ref{tab:comparison_aircraft}.
These results indicate that, under a fixed annotation budget, reallocating annotation cost from expensive fully labeled data to low-cost weakly labeled data is a key factor in achieving superior performance.

\section{Conclusion}

We proposed an active learning (AL) framework that leverages vision-language models (VLMs) as weak annotators to reduce reliance on costly human supervision.
From preliminary experiments, we found that the reliability of VLMs varied significantly with label granularity in fine-grained recognition tasks, where VLMs struggled with fine-grained labels but could provide accurate coarse-grained labels.
Motivated by this observation, the proposed AL framework combines fine-grained human annotations with coarse-grained VLM-generated weak labels through instance-wise supervision selection under a fixed annotation budget.
To mitigate the noise inherent in VLM-generated weak labels, we further incorporated a noise-robust learning strategy that models VLM-specific error patterns.
Experiments on CUB200 and FGVC-Aircraft demonstrated that the proposed method consistently outperforms conventional AL approaches under the same annotation cost, highlighting the effectiveness of integrating VLMs as cost-efficient weak annotators for fine-grained classification.

\vspace{2mm}
\noindent
\textbf{{Acknowledgements:}} This work was supported by JST ACT-X Grant Number JPMJAX23CR, JSPS KAKENHI Grant Number JP25K22846, JP22H05180, and SIP Grant Number JPJ012425.

\bibliographystyle{IEEEbib}
\bibliography{main}

\end{document}